%% file: main.tex
\documentclass[conference]{IEEEtran}
\IEEEoverridecommandlockouts
\usepackage{cite}
\usepackage{amsmath,amssymb,amsfonts}
\usepackage{algorithmic}
\usepackage{graphicx}
\usepackage{textcomp}
\usepackage{caption}
\usepackage{cuted}
\usepackage[export]{adjustbox}
\def\BibTeX{{\rm B\kern-.05em{\sc i\kern-.025em b}\kern-.08em
    T\kern-.1667em\lower.7ex\hbox{E}\kern-.125emX}}

\input{macros}

\begin{document}
\title{Scaling Bimanual Household Manipulation from 1,500 hours of Demonstrations to On-Policy Corrections
\vspace{-6mm}
\thanks{Real-robot data and UMI data are sponsored by \href{https://www.primebot.cn/}{PrimeBot} and \href{https://crobotia.com/}{crobotia} respectively.}
}

\author{
\IEEEauthorblockN{
Jiafeng Xu\textsuperscript{2},
Qi Li\textsuperscript{1},
Yan Shen\textsuperscript{2},
Yiyu Ren\textsuperscript{1},
Travis Davies\textsuperscript{1},
Shaowen He\textsuperscript{1},\\
Ze Wang\textsuperscript{3},
Yifan Yang\textsuperscript{1},
Ran Cheng\textsuperscript{1},
Hao Dong\textsuperscript{1,2}
}
\IEEEauthorblockA{
\textsuperscript{1}PrimeBot Research Institute, Swancor Advanced Materials Co., Ltd. \\
\textsuperscript{2}School of Computer Science, Peking University. \\
\textsuperscript{3}Crobotia.
\vspace{2pt}
}
\IEEEauthorblockA{
Project website: {\small\href{https://huggingface.co/datasets/challenge-2026/challenge_data}
{\textcolor{linkpink}{https://huggingface.co/datasets/challenge-2026/challenge\_data}}}
}
}

\maketitle

\begin{strip}
  \vspace*{-3.0\baselineskip}
  \centering
  \includegraphics[width=1.0\textwidth]{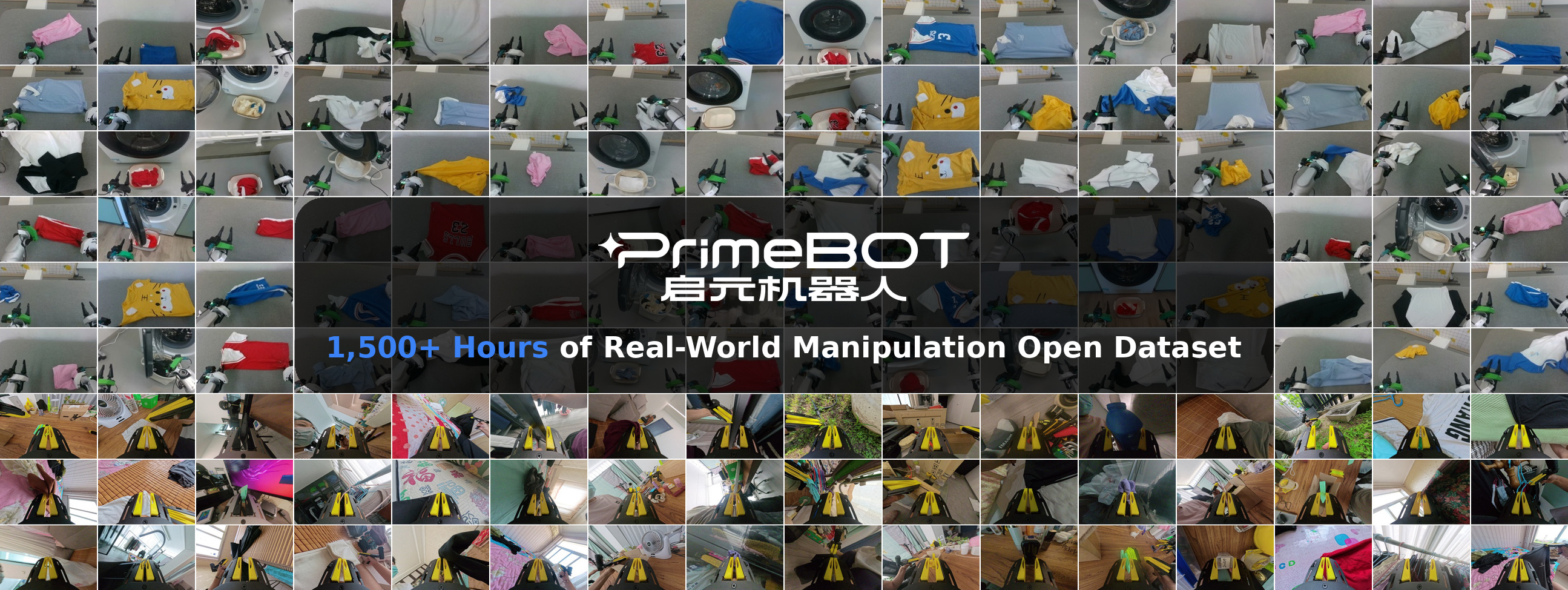}
   \captionof{figure}{
   Introducing \textbf{PrimeBot Household Manipulation Dataset}. The data encompasses core skills required for everyday housework: deformable-object manipulation, articulated-object interaction, appliance operation, and general tidying and object rearrangement. It integrates two complementary sources, bimanual real-robot teleoperation and UMI collection in diverse home scenes.}
  \label{fig:teaser}
\end{strip}

\begin{abstract}
Learning generalist policies for robust bimanual manipulation is bottlenecked by the scarcity of high quality large scale human demonstration data. In this work, we release 1,500 hours of diverse bimanual manipulation demonstrations covering everyday household tasks, and use this comprehensive corpus to train \name{}, a powerful vision‑language‑action (VLA) model. Enabled by a purpose built high throughput data pipeline and a carefully designed multi stage training paradigm, \name{} attains strong manipulation performance in our systematic experiments while retaining favorable training efficiency and high data utilization. We further study two critical scaling axes: varying the amount of expert demonstration data, and post training on DAgger correction data from real time human interventions. In both settings, task success rate improves steadily over the data ranges we probe, exhibiting a clear consistent scaling trend at our current data scale. These results validate both the learning capacity of \name{} and the promising scaling properties of the released dataset, which we open source to support reproducible research on bimanual robot manipulation learning.

\end{abstract}


\section{Introduction}
Bimanual manipulation is essential for tasks requiring two-arm coordination, long-horizon execution, and complex interactions~\cite{zhao2023learning,fu2024mobile,liu2025rdt,gkanatsios20253d,shen2026bipremanip}. Recent advances in VLA models~\cite{brohan2023rt,black2024pi0,physicalintelligence2025pi05,li2025gr} and large-scale robot datasets~\cite{o2024open,bu2025agibot,khazatsky2024droid,hou2025robomind,cheang2025gr} have made data scaling central to generalist manipulation policies~\cite{shi2026diversity,lin2025data,wang2026rethinking}. As datasets grow, however, the question shifts from how much data to collect to which data are most valuable: offline expert demonstrations provide successful behaviors, while corrective on-policy data captures deployment-time states and failures. This motivates studying how these complementary data sources scale bimanual policy learning.

\begin{figure}[!t]
    \centering
    \setlength{\abovecaptionskip}{5pt}
    \setlength{\belowcaptionskip}{-10pt}
    \captionsetup{skip=3pt}
    \includegraphics[width=1.0\linewidth]{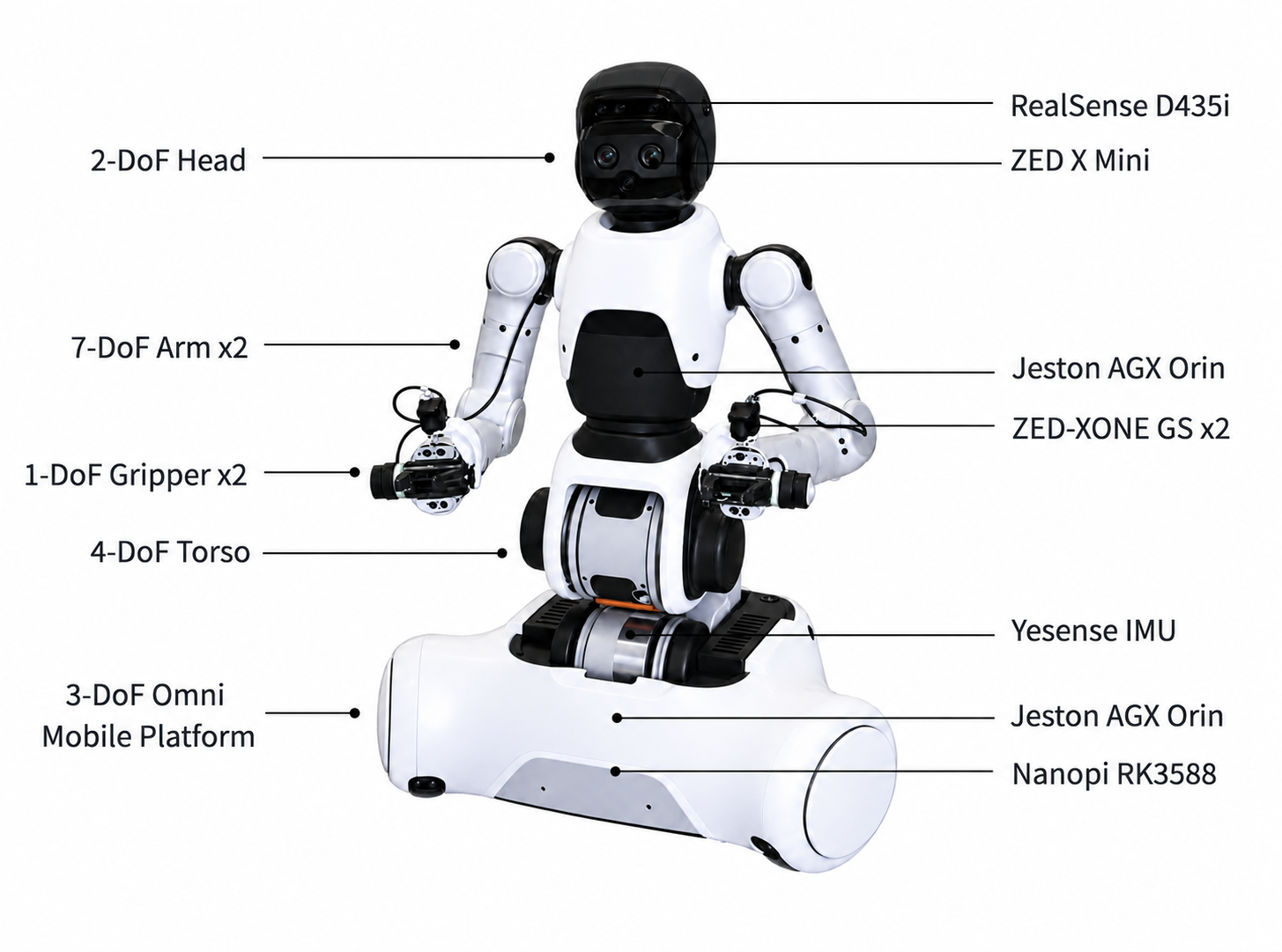}
    \caption{\hardware{} hardware configuration. The robot has 25 degrees of freedom. It is equipped with a variety of sensors, including the RealSense D435i, ZED X Mini and ZED-XONE GS cameras, as well as an inertial measurement unit (IMU) and heterogeneous computing units, supporting the performance of household bimanual manipulation tasks.}
    \label{fig:x2w_config}
\end{figure}

In this work, we first assemble a large-scale offline corpus of 1,500 hours of real-world bimanual household manipulation data, combining real-robot teleoperation and Universal Manipulation Interface (UMI) handheld-gripper demonstrations~\cite{chi2024universal}. The real-robot subset contains 32,518 trajectories, 57.4 million frames, and 531.7 hours of interaction collected using homogeneous mobile dual-arm robots. Spanning multiple household settings, the dataset covers both complete long-horizon tasks and fine-grained manipulation primitives, including garment folding, washer interaction, object transfer, and laundry-basket handling. To support learning across temporal scales, long-horizon demonstrations are aligned with sub-task-level language annotations, with semantically related instructions organized into 11 atomic skills.

Building on this corpus, we first train a VLA model from large-scale expert demonstrations, and we then use Dataset Aggregation (DAgger)~\cite{ross2011reduction} to collect human corrections during on-policy rollouts, extending supervision to policy-visited states and deployment-time failures. We aggregate these corrective trajectories with expert data for post-training, preserving skill coverage while allocating more corrective data to sub-tasks with lower success rates.

Our experiments further examine the scaling behavior of expert and corrective data. On clothes folding, performance improves as expert data increases from 10 to 120 hours, but shows little further gain at 160 hours. In contrast, DAgger targets policy-induced failure states and raises the success rate from 58\% to 93\% over three rounds. Together, these results show that expert demonstrations and corrective on-policy data play complementary roles in scaling bimanual policy learning, with corrective data providing continued gains as returns from additional expert data diminish.

\section{Robot and Dataset}

\subsection{Robot System}
We employ the \hardware{}, a mobile bimanual robotic platform designed specifically for whole-body manipulation tasks. As show in Fig.~\ref{fig:x2w_config}, the robot consists of two 7-DoF arms, each terminated by a 1-DoF gripper, together with a 4-DoF waist and a 2-DoF head, amounting to 22 controllable joints, all mounted on a three-wheel omnidirectional mobile base. The main body joints are driven by quasi-direct-drive (QDD) actuators, providing responsive and compliant motion for contact-rich bimanual manipulation, while the three independently driven wheels enable omnidirectional motion of the base within the workspace.

The proprioceptive state is a unified 89-D vector,
\begin{equation}
    \mathbf{s}_t =
    \left[
    \mathbf{q}_t,
    \dot{\mathbf{q}}_t,
    \boldsymbol{\tau}_t,
    \mathbf{p}_t^{L},
    \mathbf{p}_t^{R},
    \mathbf{q}_t^{w},
    \dot{\mathbf{q}}_t^{w},
    \boldsymbol{\tau}_t^{w}
    \right]
    \in \mathbb{R}^{89},
\end{equation}
which concatenates the positions $\mathbf{q}_t$, velocities $\dot{\mathbf{q}}_t$, and torques $\boldsymbol{\tau}_t$ of the 22
controllable joints, the poses of the two end-effectors $\mathbf{p}_t^{L,R}$, and the positions $\mathbf{q}_t^{w}$, velocities $\dot{\mathbf{q}}_t^{w}$, and torques $\boldsymbol{\tau}_t^{w}$ of the three drive wheels.

The robot’s motion is governed by 25 control variables: 22 target joint positions and 3 target wheel velocities.
\begin{equation}
    \mathbf{a}_t =
    \left[
    \mathbf{q}_t^{\mathrm{cmd}},
    \dot{\mathbf{q}}_t^{w,\mathrm{cmd}}
    \right]
    \in \mathbb{R}^{25}.
\end{equation}

For visual perception, the robot carries one head-mounted camera providing global scene context and two wrist-mounted cameras capturing close-range manipulation details, together forming complementary multi-view observations. 

All proprioceptive states, action sequences, and visual observations are temporally synchronized and recorded at a uniform frequency of 30 Hz.

\begin{figure*}[t]
    \centering
    \setlength{\abovecaptionskip}{0pt}
    \setlength{\belowcaptionskip}{-10pt}
    \includegraphics[
        width=0.95\textwidth,
        keepaspectratio
    ]{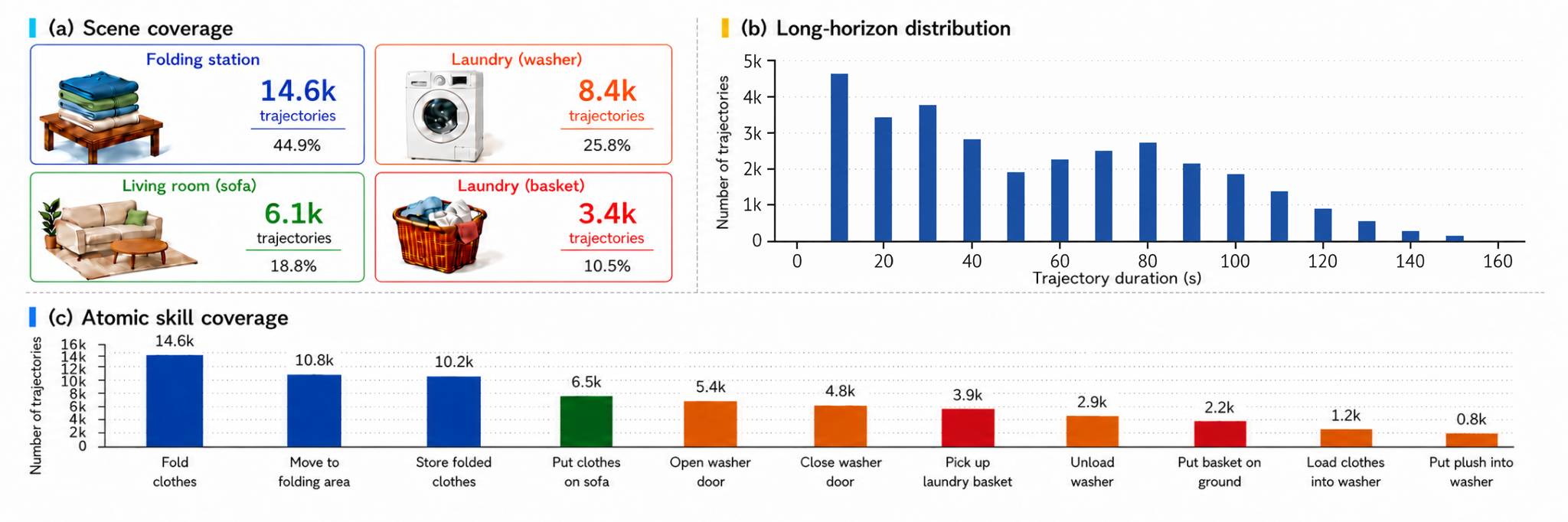}
    \caption{
        \textbf{Dataset statistics.}
        \textbf{(a) Scene coverage.} Real-robot trajectories span four household interaction settings: folding station, laundry washer, sofa, and laundry basket.
        \textbf{(b) Long-horizon distribution.} The duration distribution exhibits broad coverage of long-horizon episodes, with a long tail reaching around 150 s.
        \textbf{(c) Atomic skill coverage.} Distribution of 11 atomic skills, covering high-frequency garment manipulation as well as less frequent but task-critical object-transfer primitives.
    }
    \label{fig:dataset_overview}
\end{figure*}

\subsection{Data Collection Pipeline}
Our data collection system consists of two complementary pipelines: offline expert demonstration collection and online DAgger corrective data collection. The former builds large-scale, high-quality real-robot demonstrations through standardized teleoperation, while the latter continuously collects corrective demonstrations from human interventions during real-world policy deployment.

\paragraph{Expert Demonstration Collection.}
For offline real‑robot data collection, long‑horizon tasks are decomposed into sub‑tasks guided by their inherent structure and interaction contexts. Considering the robot’s reachable workspace and physical limitations, we design a Standard Operating Procedure (SOP) for every sub‑task to standardize scene setup, object arrangement, robot initialization, and manipulation workflows. Remote human operators use VR devices to tele‑operate the robot and record demonstration trajectories while adhering to the predefined SOPs.

Before being uploaded to the cloud, the raw multimodal trajectory data undergo automatic post-processing and format conversion. Once stored in the cloud, the data pass through automated quality validation, VLM‑powered task and language annotation, and a final manual review stage. This end‑to‑end standardized pipeline produces consistent, high‑quality demonstrations across varying robots, scenes, and operators.

\paragraph{Online DAgger Collection.}
In addition to offline expert demonstrations, we collect online corrective data during real-world policy deployment. Multiple robots execute complete long-horizon tasks using a shared policy. Whenever the policy reaches a failure state or requires correction, a human operator temporarily takes control, corrects or completes the current sub-task, and immediately returns control to the policy. As a result, a single long-horizon rollout may contain multiple alternating policy- and human-controlled segments without restarting the task after each failure.

The system automatically records all human intervention segments as corrective demonstrations, associates them with the corresponding sub-task prompts, and explicitly distinguishes policy-generated and human-controlled segments within the same trajectory. Thus, DAgger data requires no additional re‑annotation on the cloud. As the number of deployed robots increases, failures encountered during actual execution are continuously transformed into new corrective demonstrations, forming a scalable closed-loop data pipeline for improving long-horizon manipulation policies.

\subsection{Real-robot Dataset Overview}
Our real-robot consists of 32,518 teleoperated controlled real robot trajectories, covering more than 12 everyday household tasks, totaling approximately 57.4 million frames and 531.7 hours of interaction data, all with frame-level language annotations. Representative trajectories are visualized in Fig.~\ref {app:robot_view} of the appendix.

\paragraph{Scene coverage.}
As shown in Fig.~\ref{fig:dataset_overview}(a), the real-robot dataset spans four household settings: \textit{Folding Station}, \textit{Laundry (Washer)}, \textit{Living Room (Sofa)}, and \textit{Laundry (Basket)}, containing approximately 14.6k, 8.4k, 6.1k, and 3.4k trajectories, respectively. These settings cover garment manipulation, washer interaction, object transfer, and laundry-basket handling.

\paragraph{Long-horizon manipulation.}
The dataset contains both complete \textit{long-horizon tasks} and \textit{skill-level demonstrations}. As illustrated in Fig.~\ref{fig:dataset_overview}(b), trajectory durations range from short atomic primitives to long sequences exceeding 100 s, with some beyond 150 s. For multi-stage tasks, we provide language-aligned sub-step annotations, enabling supervision at both the task and skill levels.

\paragraph{Atomic skill coverage.}
After merging semantically equivalent instructions, we obtain 11 atomic skills, as summarized in Fig.~\ref{fig:dataset_overview}(c). The dataset includes both high-frequency skills, such as \textit{Fold clothes} (14.6k), and less frequent but task-critical primitives, such as \textit{Load clothes into washer} (1.2k) and \textit{Put plush into washer} (0.8k). Since a long-horizon trajectory may contain multiple skills, skill counts are not mutually exclusive. Overall, the dataset provides unified multi-scene coverage, long-horizon supervision, and fine-grained skill annotations for household manipulation.

\subsection{UMI Dataset Overview}
Complementing the teleoperated corpus, we collect approximately 1{,}000 hours of in-the-wild bimanual UMI demonstrations of household garment manipulation (folding and organizing clothes), matching the task distribution of the \textit{Folding Station} setting. Each demonstration provides synchronized dual egocentric video ($960\times960$ at 30\,Hz), metric 6-DoF end-effector trajectories recovered by visual-inertial SLAM, continuous gripper aperture, and frame-level language annotations at two granularities: item-level segments (e.g., \textit{fold the red shirt}) and fine-grained action steps within each segment (e.g., \textit{grasp}, \textit{lay flat}, \textit{fold one side over}). 

\paragraph{In-the-wild diversity.}
Freed from a fixed workcell, the data are collected in more than 200 households by over 100 collectors, spanning 5{,}000 distinct garments that vary in category, fabric, color, and print. Demonstrations take place on beds, tabletops, and drying racks, and are recorded at all hours, from morning to late night, so lighting ranges from bright daylight to dusk and artificial illumination. Folding strategies vary across collectors as well. The resulting coverage of scenes, objects, illumination, and manipulation styles exceeds what a fixed robot cell can afford, representative trajectories are shown in the appendix (see Fig.~\ref{app:umi_view}).

\begin{figure*}[t]
    \centering
    \setlength{\belowcaptionskip}{-10pt}
    \includegraphics[width=1.0\textwidth]{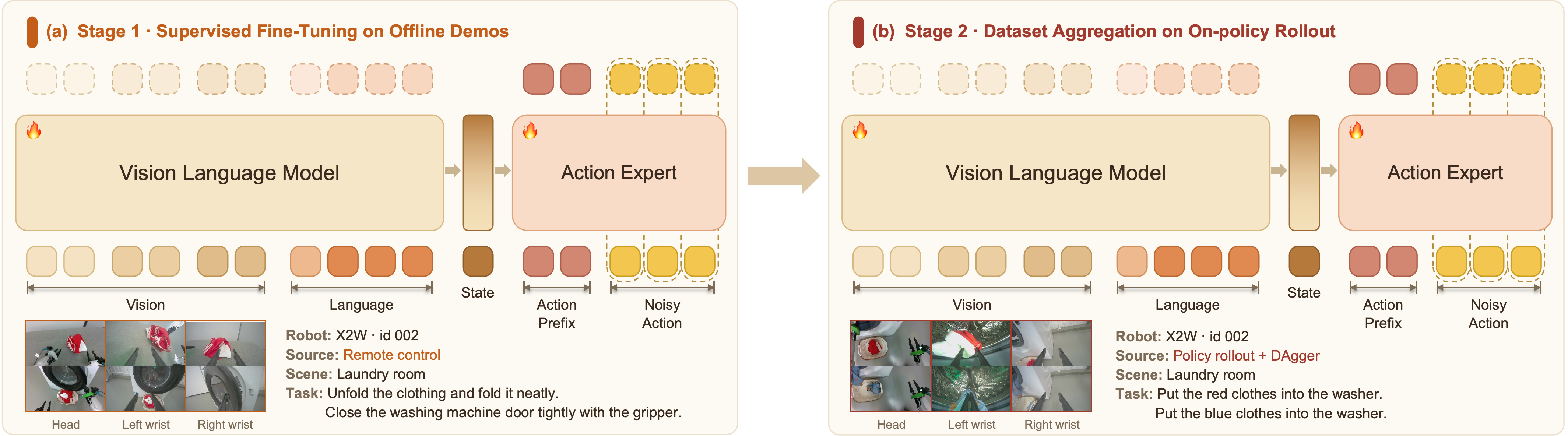}
    \caption{\textbf{Two-stage full-parameter training of \name{}.} Both stages use the same VLM + action-expert architecture with all parameters trainable (flames); dashed tokens are model outputs. \textbf{(a)} Stage~1 fits offline teleoperated demonstrations $\mathcal{D}_{\mathrm{demo}}$ on expert-visited states.
    \textbf{(b)} Stage~2 initializes from Stage~1, rolls out the policy, collects expert corrections $a_{\mathrm{expert}}$ on the visited states, and retrains on the aggregated data---aligning training with the policy's own state distribution.}
    \label{fig:model}
\end{figure*}

\paragraph{Bimanual, encoder-free capture.}
Both grippers are recorded simultaneously, yielding synchronized dual-view, dual-trajectory demonstrations of intrinsically two-handed garment manipulation; in-the-wild bimanual data at this scale remains scarce. Each gripper is fully mechanical and self-contained, following a vision-only design: the wrist-mounted camera is its sole sensor, the aperture is recovered visually rather than from encoders, and no head-mounted or third-person camera is involved. Extensive policy-learning experiments validate this design choice: wrist-view-only observation is sufficient for bimanual coordination in manipulation tasks~\cite{chi2024universal}, hand-centric views improve training efficiency and out-of-distribution generalization over third-person views~\cite{hsu2022vision}, and recent in-the-wild systems with hand-centric-only sensing achieve zero-shot deployment in unseen homes~\cite{etukuru2025robot}; head-mounted views are instead primarily useful for navigation and mobile-base control rather than tabletop manipulation. 

The gripper is also built for endurance. Free of encoders, motors, and actuation electronics, it is markedly lighter than instrumented alternatives, allowing operators to comfortably collect data for hours at a time. Its low‑power design supports full‑day standby and ultra‑long task capture, enabling reset‑free sessions of up to 90 minutes of continuous recording. Meanwhile, the directly hand‑driven jaws deliver strong yet finely modulated grasp forces, sufficient to tension and pin fabric during folding.

\section{Model and Training Recipe}
\subsection{Main Architecture}
\name{} is a VLA model built upon a Mixture-of-Transformer architecture with a total of 5B parameters (Fig.~\ref{fig:model}), consisting of a pre-trained vision-language backbone, \textit{i.e.}, Qwen3-VL-4B-Instruct~\cite{bai2025qwen3} for multimodal perception, and an action diffusion transformer trained by flow matching~\cite{lipman2022flow} objectives as an action expert. For the control task, \name{} generates a ${K}$-length action chunk $\ba_t  = a_{t:t+K}$ conditioned on the vision observations $\mathbf{o}_{t}$, language $l$ and proprioceptive state $\mathbf{s}_{t}$, \textit{i.e.}, $\mathbf{a}_{t} = \pi_{\theta}(\mathbf{o}_{t}, \mathbf{s}_{t}, l)$.

Specifically, the action expert is constructed from a stack of 18 transformer blocks with hidden dimension \(D_{\text{hidden}} = 1024\) with 8 attention heads, which employ the grouped key-value attention~\cite{ainslie2023gqatraininggeneralizedmultiquery} mechanism with 4 grouped KV heads. Within each block, the noisy action tokens first undergo bidirectional self-attention, then cross-attend to the backbone's hidden states, and finally pass through a SwiGLU feed-forward layer. The cross-attention is depth-aligned: block $i$ attends to the $i$-th of the 18 exposed backbone layers, so the expert progressively grounds its predictions in increasingly abstract VL features rather than collapsing all conditioning onto the final layer. The action expert is further conditioned on denoising-timestep information supplied through adaLN-zero modulation.

To facilitate seamless asynchronous execution, we further train the expert module with action prefix conditioning (on the right of Fig.~\ref{fig:model}). The delay $d$ defines a leading action segment regarded as previously committed; the corresponding tokens are anchored to their ground-truth values with their timesteps fixed to \(t=1\), while only the subsequent suffix sequence undergoes noise injection, denoising, and loss supervision. During inference, the delay $d$ is dynamically computed as the average execution duration of the most recent $R$ inference rounds. Full action sequences are synthesized through five steps of Euler forward integration, supporting low-latency, smooth real-time robotic control.

\subsection{Post-training on Expert Data}
After collecting expert demonstrations, we train \name{} using the conditional flow matching objective~\cite{lipman2024flowmatchingguidecode}, enabling a single policy to learn multiple sub-tasks jointly:
\begin{equation}
    \mathcal{L}_\mathrm{CFM}(\theta) =
    \mathbb{E}_{\{\ba, \bo, \bs, l\}\sim \mathcal{D}, \epsilon, t}
    \left\|
    \pi_{\theta}(\mathbf{a}_{t}, \mathbf{o}, \mathbf{s}, l)
    - (\mathbf{a}-\epsilon) 
    \right\|^{2},
\end{equation}
where $\epsilon \sim \mathcal{N}(\mathbf{0}, \mathbf{I})$ and
$t \sim \mathrm{Beta}(1.0, 1.5)$.
The noisy action chunk is constructed as $\mathbf{a}_{t} = (1-t)\epsilon + t\mathbf{a}$.
The policy $\pi_{\theta}$ predicts the target flow $(\mathbf{a}-\epsilon)$ conditioned on the noisy action, visual observations $\mathbf{o}$, robot state $\mathbf{s}$, and language instruction $l$.

\subsection{Post-training on DAgger Data}
The collected DAgger trajectories are streamed to the cloud, where the policy is further optimized for failure cases in test environment. This stage poses two challenges: the policy must acquire failure recovery from the online corrections without forgetting the skills it already has, and sub-tasks with different learning dynamics must be trained in a balanced manner. We address both through data allocation.

First, we mix DAgger and expert data at a $1{:}1$ ratio in the number of \emph{trajectories} rather than frames. Because trajectory lengths differ across sub-tasks, frame-level mixing allows random sampling to distort skill coverage, whereas matching trajectory counts preserves the expert skill distribution.

Second, we bias the DAgger budget toward sub-tasks the policy handles poorly. Let $N$ denote the number of sub-tasks and let $s_i \in [0,1]$ be the success rate of the current policy on sub-task $i$, measured on a held-out evaluation set. The fraction $w_i$ of DAgger trajectories allocated to sub-task $i$ is

\begin{equation}
  w_i = w_{\min} + \bigl(1 - N w_{\min}\bigr)\,
        \frac{\bigl(1 - s_i + \epsilon\bigr)^{\alpha}}
             {\sum_{j=1}^{N} \bigl(1 - s_j + \epsilon\bigr)^{\alpha}},
  \label{eq:dagger-alloc}
\end{equation}

\noindent where $1 - s_i$ is the demand signal, $\alpha \geq 0$ controls how sharply the budget concentrates on weak sub-tasks ($\alpha = 0$ recovers a uniform allocation), $\epsilon > 0$ keeps $w_i$ positive once a sub-task is solved, and $w_{\min}$ floors every share; we set $w_{\min} = 0.05$, $\epsilon = 0.05$, and $\alpha = 1$. Reserving $N w_{\min}$ before the proportional split keeps the allocation closed-form, with $\sum_{i=1}^{N} w_i = 1$ and $w_i \geq w_{\min}$. The floor also guards against forgetting: strong sub-tasks retain data as the budget shifts toward the weak ones.

\section{Experiment}

\subsection{Experiment Settings}
We deploy the learned policy on an NVIDIA GeForce RTX 4090 GPU. The policy performs asynchronous inference at a frequency of 10 Hz, and generates an action chunk $\ba_t \in \mathbb{R}^{50\times 25}$. After temporal ensembling, the whole-body joint commands are synchronously issued at a fixed frequency of 30 Hz. The commands are then optimized in real-time by a whole-body motion controller derived from the teleoperation phase, controlling the robot to perform movements at a frequency of 1000 Hz.

The test environments in this experiment all incorporated slight generalizations, including variations in material size, material color, and initial robot position. After detailed testing, we found that a model achieving an 85\% success rate under generalized conditions can achieve a success rate exceeding 95\% under in-distribution conditions.

\begin{figure}[!t]
    \centering
    \setlength{\belowcaptionskip}{-10pt}
    \captionsetup{skip=3pt}
    \hspace{-0.03\linewidth}\includegraphics[width=1.0\linewidth]{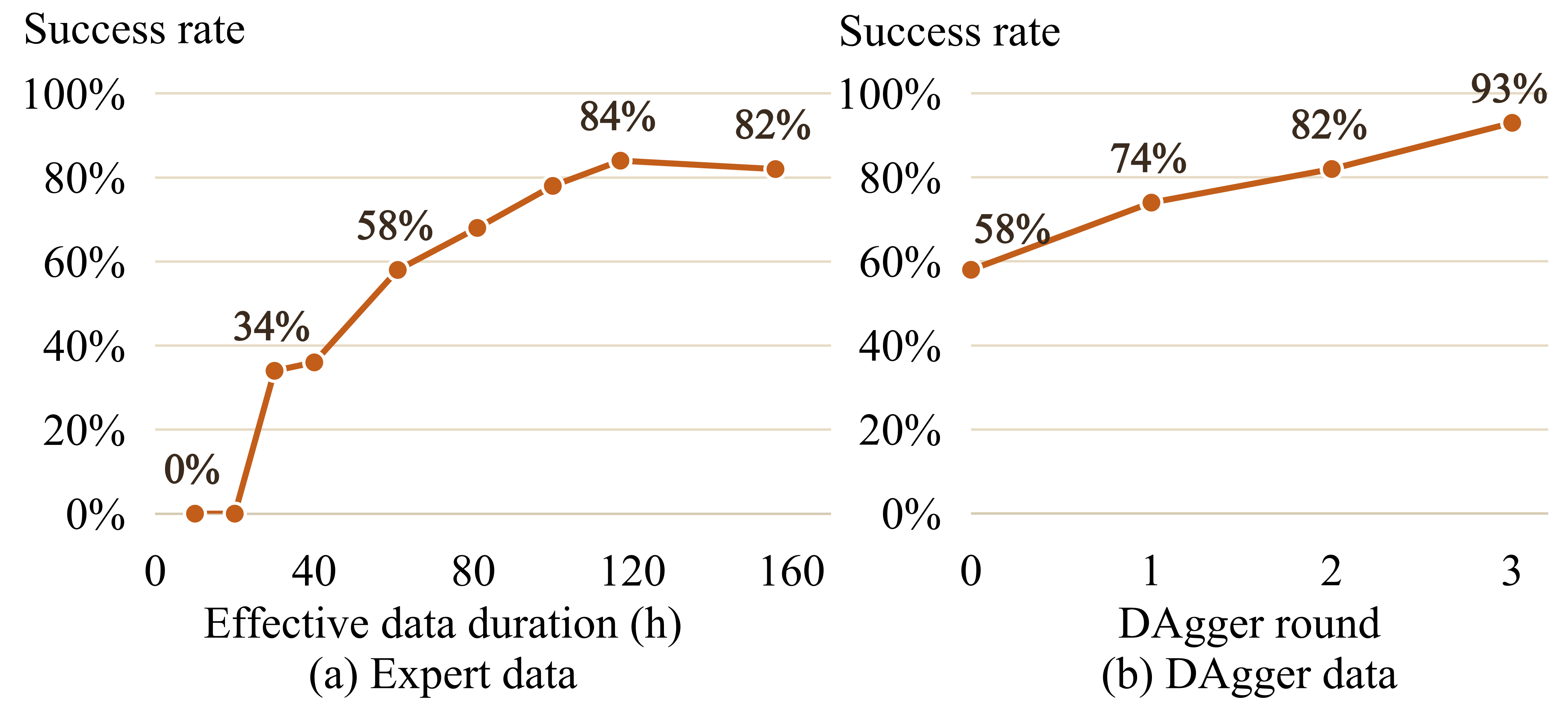}
    \caption{Data scaling in the folding-clothes task.
    (a) Success rate versus the duration of effective expert data.
    (b) Success rate over three rounds of DAgger training initialized from a
    checkpoint with a 58\% success rate.}
    \label{fig:exp}
\end{figure}

\subsection{Scaling of Expert Data}
To assess data scalability, we study folding clothes, and scale the training set from 10 hours to 160 hours (8,000 trajectories). As shown in Fig.~\ref{fig:exp}(a), the success rate monotonically increases from 34\% at 30 hours to 84\% at 120 hours, then tends to saturate, and the last 40 hours do not bring further improvement (it is 82\% at 160 hours). The scale of the released dataset is what makes this transition measurable: it locates the point at which expert demonstrations already cover the task distribution densely. Beyond it, the residual failures come from states the policy reaches only at test time, which lie outside the expert distribution and cannot be supplied by more expert data. Closing this gap requires human-in-the-loop intervention data.

\subsection{DAgger Experiment}
Given the time overhead associated with data collection and constraints on computational and experimental resources, we initialize our DAgger experiments using a checkpoint trained on 18,695 expert trajectories spanning 10 sub‑tasks. For the clothes‑folding task, the DAgger trajectories for each round are calculated according to Eq.~\ref{eq:dagger-alloc}. Each iteration conducts 2.5 epochs of post‑training, and performance improves steadily, as shown in Fig.~\ref{fig:exp}(b): the success rises from 58\% to 74\%, 82\%, and 93\% following the first, second, and third iterations, respectively, representing a 35‑point gain compared with the expert‑only baseline.

\section{Conclusion}
We release 1{,}500 hours of bimanual household manipulation data, 531.7 hours of real-robot teleoperation across 32{,}518 trajectories together with roughly 1{,}000 hours of UMI demonstrations, and train \name{}, a 5B VLA model, on it. On clothes folding, the success rate rises monotonically with the amount of expert data across the range we probe, from 34\% at 30 hours to 84\% at 120 hours, showing that the released corpus supports sustained scaling; past that point, three rounds of DAgger post-training with a failure-weighted budget raise the success rate from 58\% to 93\%. Scaling data and structuring it around on-policy failures are therefore complementary levers rather than competing ones. The complete corpus, together with its temporally aligned sub-task language annotations, is released to the community as a shared basis for studying data-centric scaling in bimanual manipulation.

\section*{Acknowledgment}
We thank the data team for building the data pipeline system, which enables efficient and rapid data accumulation and supports agile iteration; we also thank the data acquisition team for demonstrating outstanding operational skills. We thank crobotia for providing a large amount of UMI data; and we also thank the open source community for its valuable feedback and active participation throughout this project.

\clearpage
\bibliographystyle{IEEEtran}
\bibliography{reference}

\clearpage
\appendix

\section*{Representative Trajectories}
\begin{figure*}[!p]
  \centering
  \includegraphics[width=\textwidth]{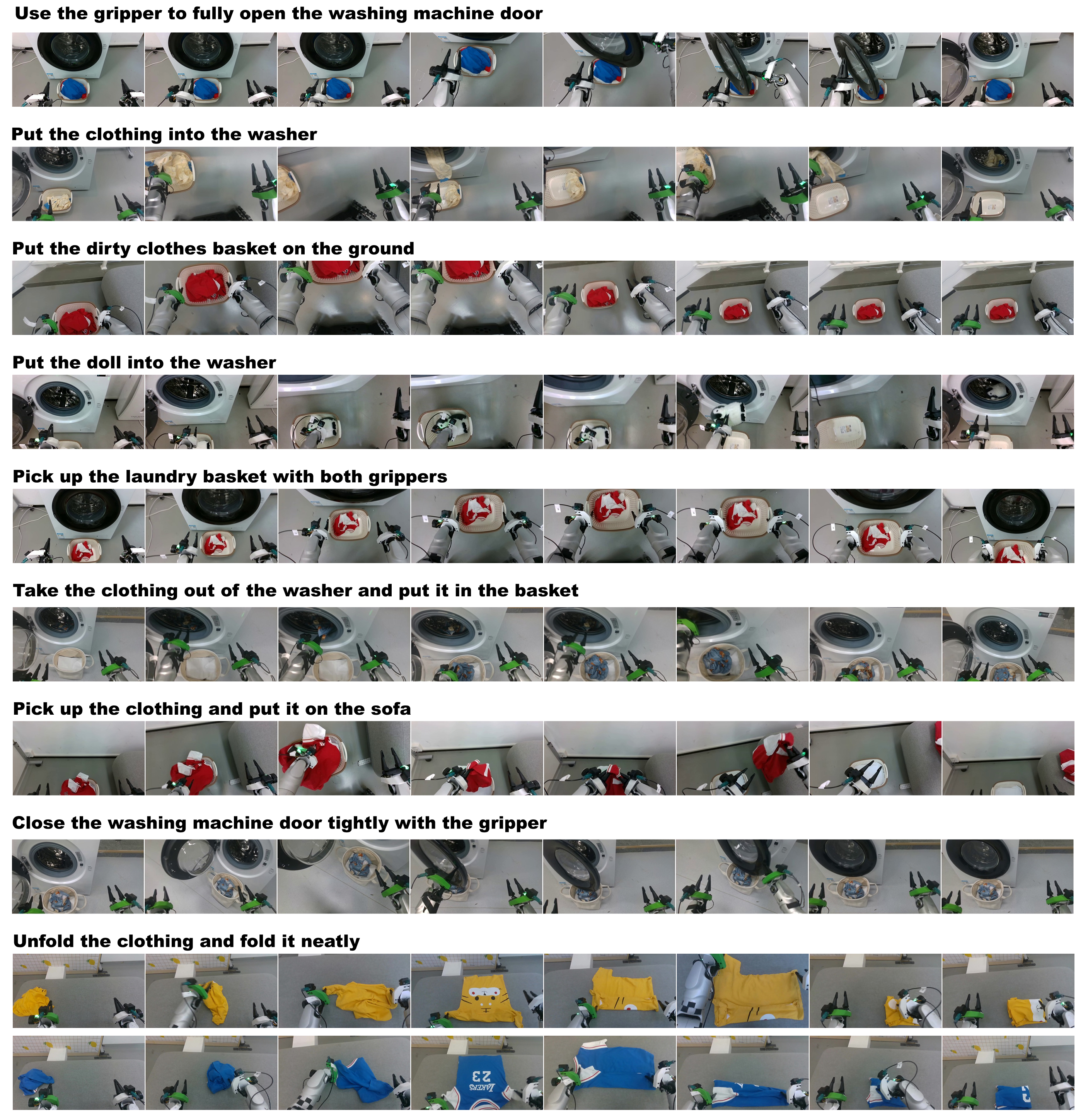}
  \caption{Representative real-robot demonstrations from the robot-mounted head camera. Each row shows a single trajectory: the heading above each image strip denotes the corresponding subtask instruction, and several frames are sampled along the trajectory to illustrate the progression of task execution. The examples cover a diverse set of household manipulation tasks, including washing-machine interaction, garment loading and unloading, object placement, bimanual laundry-basket manipulation, garment transfer, and long-horizon garment unfolding and folding.}
  \label{app:robot_view}
\end{figure*}

\begin{figure*}[!p]
  \centering
  \includegraphics[width=\textwidth]{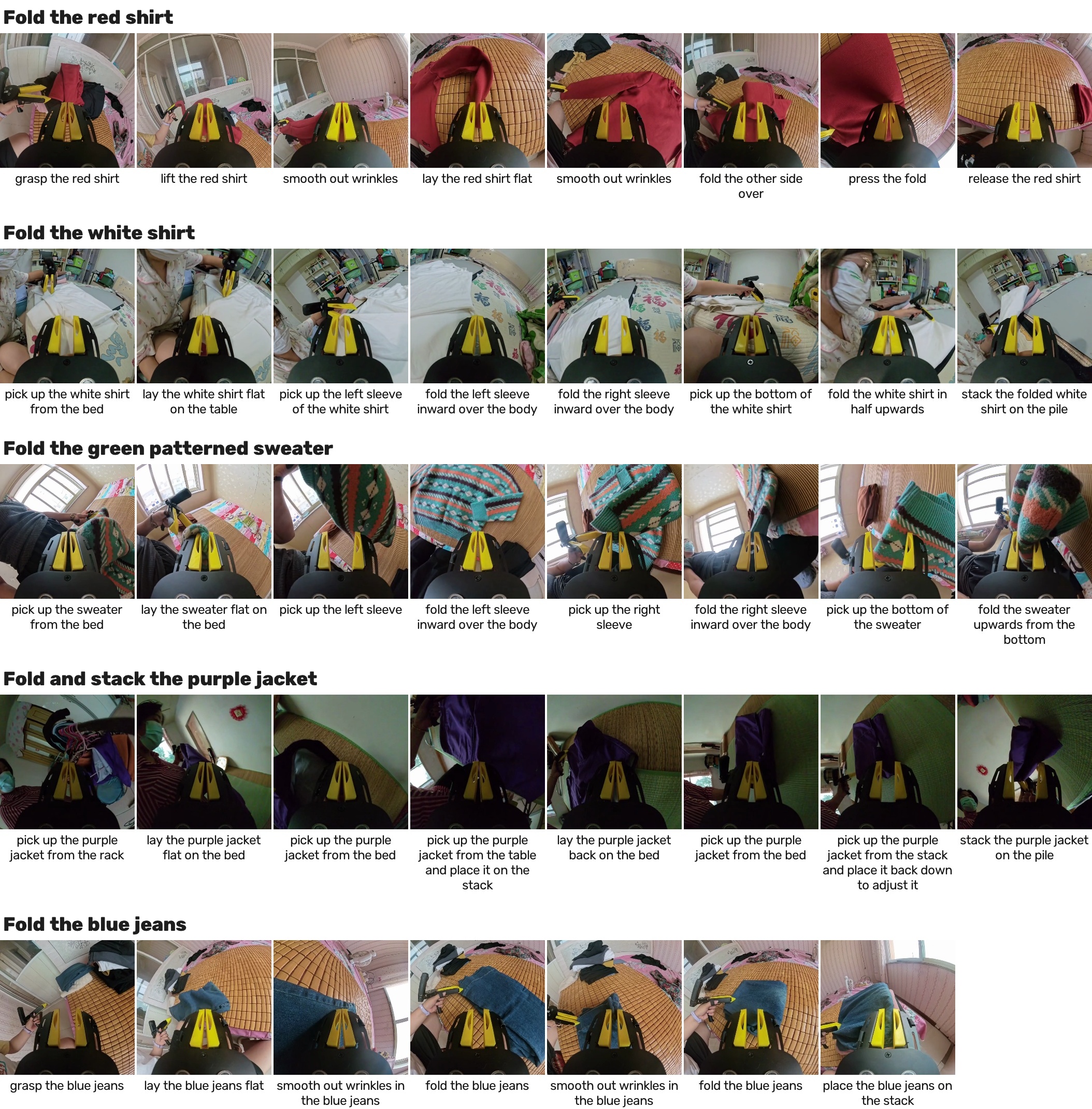}
  \caption{Representative UMI demonstrations from the egocentric wrist-mounted view. Each row shows one trajectory: the heading above each strip is the item-level task instruction, frames are sampled at the midpoints of the annotated fine-grained action steps, and the caption under each frame is the corresponding action-step annotation. Rows span four distinct households and five garment types, covering flat folding, sleeve-first folding, bimanual coordination, rack retrieval, and stacking onto a pile.}
  \label{app:umi_view}
\end{figure*}

Fig.~\ref{app:robot_view} and \ref{app:umi_view} show representative trajectories from the released real‑robot and UMI datasets, respectively. Each row is a single episode sampled at a fixed interval, with the sub-task instruction active at that time step shown above the frames.

\section*{Robot Description}
We make the robot hardware description publicly available to facilitate non‑commercial projects for teaching, experimental work, and academic research. The corresponding URDF files can be found in the \href{https://huggingface.co/datasets/challenge-2026/challenge_data/tree/main/robot_description}{\textcolor{blue}{challenge\_data}} repository.

\section*{Change Log}
\paragraph{September 2026}: The challenge is in full swing, welcome to join now!

\paragraph{August 2026}: Submit an workshop article to IROS.

\paragraph{July 2026}: Release full data and launch the \href{https://bimanual-robot-learning.github.io/challenge/}{\textcolor{blue}{IROS-2026 Household Bimanual Manipulation Challenge}}.

\end{document}

%% file: macros.tex
\usepackage{url}
\usepackage{amssymb}
\usepackage[utf8]{inputenc}
\usepackage{microtype}
\usepackage{booktabs}
\usepackage{pifont} 
\usepackage{multirow}
\usepackage{makecell}
\usepackage{paralist}
\usepackage{xspace}
\usepackage{color}
\usepackage{xcolor}
\usepackage{colortbl}
\usepackage{adjustbox}
\usepackage{hyperref} 
\usepackage[edges]{forest}
\usepackage{tikz} 
\usepackage{amsfonts}
\usepackage{bbm}
\usepackage{enumitem}
\usepackage{mathtools}
\usepackage{subcaption}

\definecolor{linkpink}{RGB}{200, 50, 150}

\newcommand{\name}{XR-2}
\newcommand{\hardware}{X2W}
\newcommand{\bo}{\mathbf{o}}
\newcommand{\ba}{\mathbf{a}}
\newcommand{\bs}{\mathbf{s}}

\newcommand{\commenthere}[1]{}

\renewcommand{\paragraph}[1]{\vspace{0.1em}\noindent\textbf{#1}}